%% file: _main.tex
\documentclass[sigconf,nonacm]{aamas}

\usepackage{balance} 

\usepackage{hyperref}       

\usepackage[ruled,vlined,noend]{algorithm2e}
\usepackage{amsmath}
\usepackage{multirow}
\usepackage{makecell}
\usepackage{nicematrix}
\usepackage{amsfonts}
\usepackage{tikz}
\usepackage[normalem]{ulem}
\usetikzlibrary{positioning}

\usepackage{algpseudocode}

\setcopyright{ifaamas}
\acmConference[AAMAS '23]{Proc.\@ of the 22nd International Conference
on Autonomous Agents and Multiagent Systems (AAMAS 2023)}{May 29 -- June 2, 2023}
{London, United Kingdom}{A.~Ricci, W.~Yeoh, N.~Agmon, B.~An (eds.)}
\copyrightyear{2023}
\acmYear{2023}
\acmDOI{}
\acmPrice{}
\acmISBN{}

\acmSubmissionID{15}

\title[WorldCloner]{Neuro-Symbolic World Models for Adapting to Open World Novelty}

\author{Jonathan C. Balloch}
\affiliation{
  \institution{College of Computing \\
  Georgia Institute of Technology}
  \city{Atlanta}
  \country{USA}}
\email{balloch@gatech.edu}

\author{Zhiyu Lin}
\affiliation{
  \institution{College of Computing \\
  Georgia Institute of Technology}
  \city{Atlanta}
  \country{USA}}
\email{zhiyulin@gatech.edu}

\author{Robert Wright}
\affiliation{
  \institution{Georgia Technology \\
  Research Institute}
  \city{Atlanta}
  \country{USA}}
\email{robert.wright@gtri.gatech.edu}

\author{Xiangyu Peng}
\affiliation{
  \institution{College of Computing \\
  Georgia Institute of Technology}
  \city{Atlanta}
  \country{USA}}
\email{xpeng62@gatech.edu}

\author{Mustafa Hussain}
\affiliation{
  \institution{College of Computing \\
  Georgia Institute of Technology}
  \city{Atlanta}
  \country{USA}}
\email{mustafa.hussain@gatech.edu}

\author{Aarun Srinivas}
\affiliation{
  \institution{College of Computing \\
  Georgia Institute of Technology}
  \city{Atlanta}
  \country{USA}}
\email{asrinivas5@gatech.edu}

\author{Julia M. Kim}
\affiliation{
  \institution{College of Computing \\
  Georgia Institute of Technology}
  \city{Atlanta}
  \country{USA}}
\email{julia.kim@gatech.edu}

\author{Mark O. Riedl}
\affiliation{
  \institution{College of Computing \\
  Georgia Institute of Technology}
  \city{Atlanta}
  \country{USA}}
\email{riedl@cc.gatech.edu}

\begin{abstract}
Open-world \textit{novelty}--a sudden change in the mechanics or properties of an environment–is a common occurrence in the real world. 
\textit{Novelty adaptation} is an agent’s ability to improve its policy performance post-novelty. 
Most reinforcement learning (RL) methods assume that the world is a closed, fixed process. 
Consequentially, RL policies adapt inefficiently to novelties. 
To address this, we introduce \sysname{}, an end-to-end trainable neuro-symbolic world model for rapid novelty adaptation. 
\sysname{} learns an efficient symbolic representation of the pre-novelty environment transitions, and uses this world model to detect novelty and efficiently adapt to novelty in a single-shot fashion. 
Additionally, \sysname{} augments the policy learning process using \textit{imagination-based adaptation}, where the world model simulates transitions of the post-novelty environment to help the policy adapt. 
By blending ``imagined'' with interactions in the post-novelty environment, performance can be recovered with fewer total environment interactions. 
Using environments designed for studying novelty in sequential decision-making problems, we show that the symbolic world model helps its policy adapt more efficiently than neural-only reinforcement learning methods.
\end{abstract}


\newcommand{\sysname}{{\sc WorldCloner}}

\makeatletter
  \newcommand\reduline{\bgroup\markoverwith{\textcolor{red}{\rule[-0.5ex]{2pt}{0.4pt}}}\ULon}
  \UL@protected\def\blueuwave{\leavevmode \bgroup 
    \ifdim \ULdepth=\maxdimen \ULdepth 3.5\p@
    \else \advance\ULdepth2\p@ 
    \fi \markoverwith{\lower\ULdepth\hbox{\textcolor{blue}{\sixly \char58}}}\ULon}
  \UL@protected\def\yellowdotuline{\leavevmode \bgroup 
    \UL@setULdepth
    \ifx\UL@on\UL@onin \advance\ULdepth2\p@\fi
    \markoverwith{\begingroup
       \lower\ULdepth\hbox{\kern.06em \textcolor{yellow}{.}\kern.04em}%
       \endgroup}%
    \ULon}
  \UL@protected\def\greendashuline{\leavevmode \bgroup 
    \UL@setULdepth
    \ifx\UL@on\UL@onin \advance\ULdepth2\p@\fi
    \markoverwith{\kern.13em
    \vtop{\color{green}\kern\ULdepth \hrule width .3em}%
    \kern.13em}\ULon}
\makeatother

\begingroup
\definecolor{rulecolor}{rgb}{1.0, 0.95, 0.95}

\begin{document}


\pagestyle{fancy}
\fancyhead{}


\maketitle 


\section{Introduction}

\input{1_intro}
\section{Background}
\input{2_background}
\section{WorldCloner}
\input{4_method}

\section{Experiments}
\label{sec:experiments}
\input{5_experiments}
\input{6_results}

\section{Conclusions}\label{sec:discussion}
\input{7_discussion_future}

\bibliographystyle{ACM-Reference-Format} 
\bibliography{references}


\end{document}

%% file: 1_intro.tex
\begin{figure}[t]
    \centering
    \includegraphics[width=0.9\linewidth]{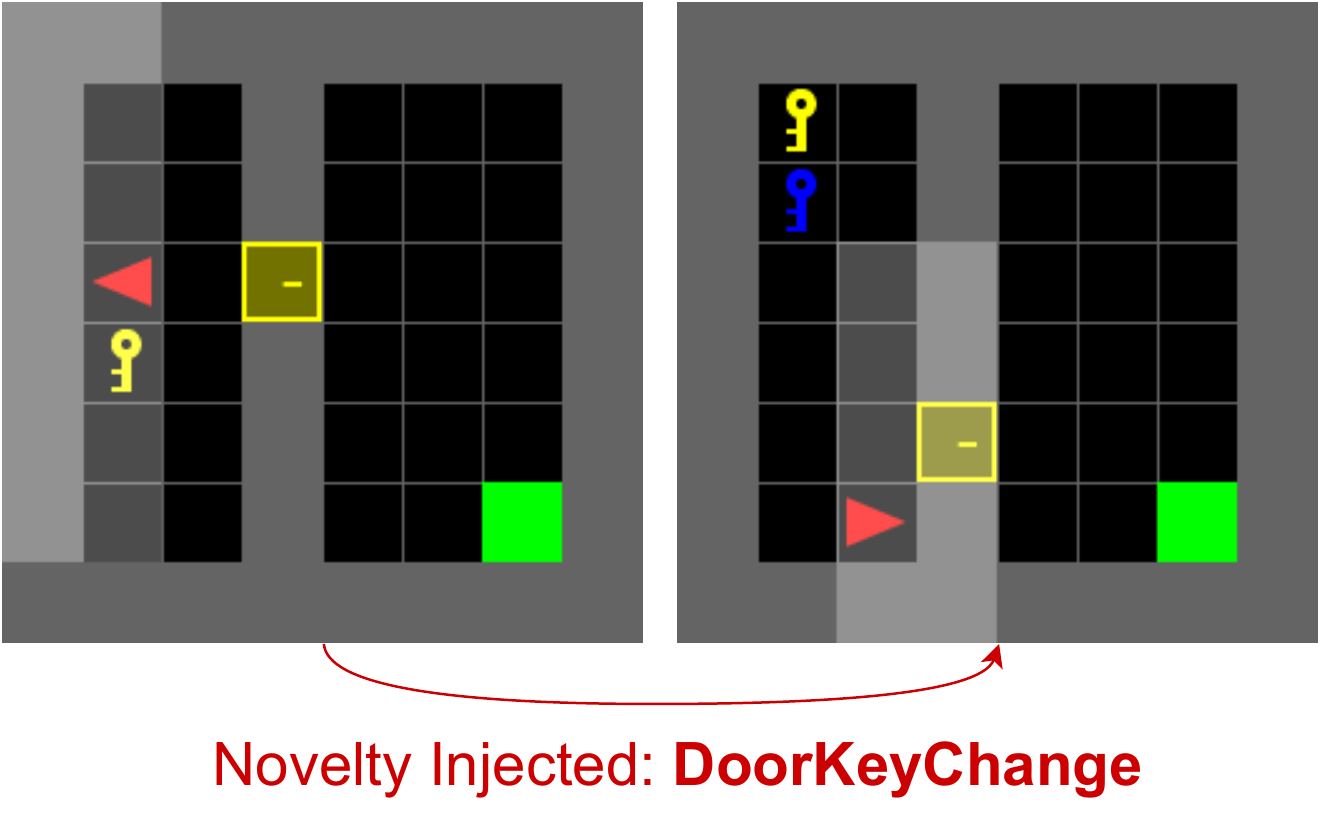}
    \includegraphics[width=0.9\linewidth]{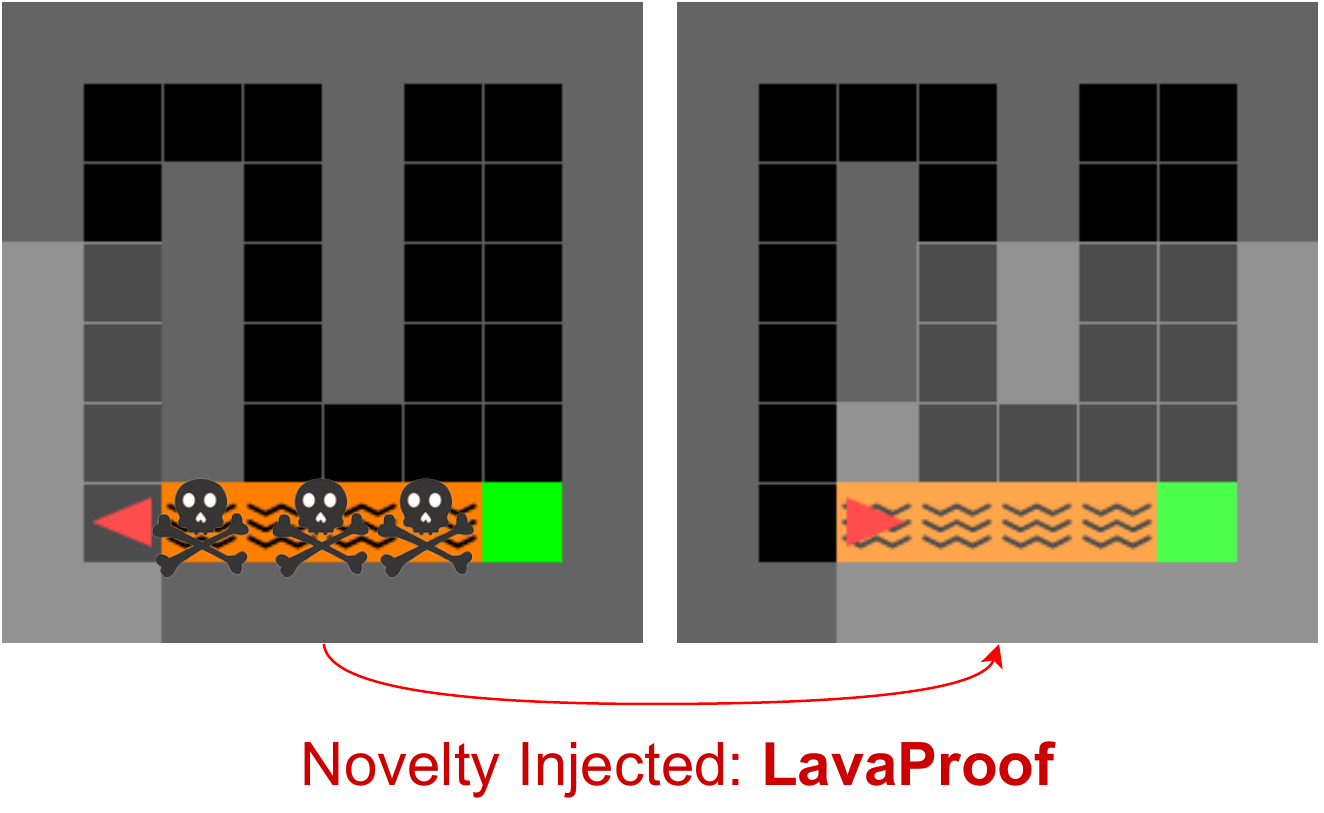}
    \caption{The NovGrid environments, where the agent (red triangle) must get to the goal (green box). The novelties are not directly observable; the agent must experience the novelty to be aware of it. 
    {\bf Top:} 
    pre-novelty only a yellow key opens a door; post-novelty only the blue key opens the door.
    {\bf Bottom:}
    pre-novelty the lava gives a -1 reward and is a terminal state; post-novelty the lava is safe to walk on.
    }
    \label{fig:splash}
\end{figure}

Most realistic environments are ``open-worlds,'' where new and unexpected situations will likely be encountered.
Consider the following examples:
a door is re-keyed so previous keys fail to open it and a different key is required, changing the solution;
a new technology enables a vehicle to pass over previously impassible terrain, shortening the distance to a destination;
a volcano eruption suddenly changes the local climate causing native crops to fail, making it harder to feed communities. 
We refer to these changes as ``\textit{novelties}'', previously unseen changes to the fundamental mechanics or properties of an environment \citep{pimentel2014review,boult2021towards}. 
Novelties can dramatically alter the solution distribution for sequential decision-making tasks.
An agent---human or artificial--- that fails to adapt to novelties can result in a precipitous drop in task performance if the same policies continue to be pursued.


\begin{figure*}[t]
    \centering
    \includegraphics[width=0.90\linewidth]{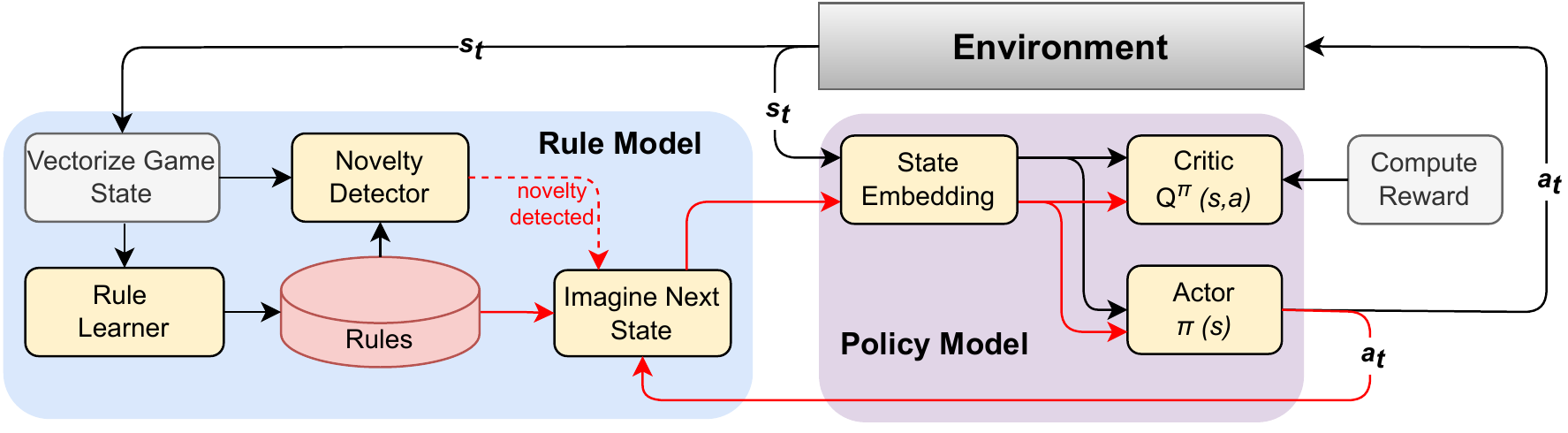}
    \caption{The \sysname{} architecture.
    The purple module 
    and black arcs represent the conventional RL execution loop with loss back-propagating backward through black arcs in the purple module.
The blue module 
contains rule model learning and novelty detection.
The red arcs represent information flow in a post-novelty environment, using learned rules to simulate the new environment.
Post-novelty, loss is back-propagated backward along the red arcs and black arcs within the policy model.}
    \label{fig:architecture}
\end{figure*}

Fast and sample-efficient adaptation to novelties can be essential, for example when human safety is involved. 
As such, humans have evolved to be good at adapting to novelties \citep{braun2009learning}. 
By contrast, deep reinforcement learning (RL) agents, which have demonstrated super-human performance on game tasks like Go~\citep{silver2017mastering}, Atari~\citep{badia2020agent57}, and StarCraft~\cite{vinyals2019grandmaster} are particularly susceptible to failure in the face of novelty.
Novelties constitute a distributional shift in the environment dynamics.
As such novelties can be expressed as online task transfer~\cite{balloch2022role}---during online execution the agent suddenly finds itself in an environment with new transition dynamics.
The traditional approach to transfer is to retrain from scratch in the post-novelty environment. 
One can also allow the agent to continue learning post-novelty starting with the model parameters.  
This, however, can be very sample-inefficient, requiring hundreds of thousands of interactions with the environment during deployment time to converge to the new optimum policy.

We propose \sysname, an efficient \textit{world model} reinforcement learning system with a neural policy comprised of two novelty-focused improvements to the standard deep RL execution loop.
(1)~We propose a simple, fast-updating, symbolic world model for learning a model of the transition function---how features of the environment change and can be changed over time.
Unlike neural world models which may require multiple examples of a phenomenon, our symbolic world model can be updated with a single post-novelty observation, allowing faster novelty adaptation than neural world models.
(2)~We propose an {\em imagination-based adaptation} method that uses the symbolic world model to improve the efficiency of deployment-time adaptation of the policy. 
After detecting novelty, the agent updates its world model and uses the updated world model to simulate environment transitions in the post-novelty world, reducing the number of real environment interactions required to update the policy.

We evaluate the sample efficiency of \sysname{} in the NovGrid environment~\citep{balloch2022novgrid} with multiple novelty types.
We show that post-novelty adaptation with \sysname{} requires fewer policy updates and environment interactions
than model-free and neural world model reinforcement learning techniques.
%

%
To summarize, our contributions are as follows:
\begin{itemize}
    \item We present \sysname, a neuro-symbolic world model for novelty detection and adaptation.
    \item We define a new symbolic representation with an efficient learning algorithm and a way to use this representation to help world models adapt to novelty.
    \item We show that~\sysname{} adapts to novelties more efficiently than state-of-the-art reinforcement learners.
\end{itemize}

%% file: 2_background.tex
\paragraph{\bf Novelty-Handling.}
Novelties are defined as sudden, previously unseen changes to dataset, datastream, or environment fundamentals \citep{pimentel2014review,boult2021towards}.
Novelty handling can be broken down into three challenges: {\em novelty detection}, {\em novelty characterization}, and {\em novelty adaptation}.
Prior work on novelty-aware agents for sequential decision making problems include
adaptive mixed continuous-discrete planning and knowledge graphs used in combination with actor-critic reinforcement learning techniques to improve both detection and adaptation \citep{klenk2020model,peng2021detecting,sarathy2021spotter,loyall2021integrated}. 
Adaptation to ``hidden'' domain shifts~\citep{chen2021active} is also relevant, though not originally targeted toward novelty.

Explicit detection and characterization are not strictly necessary for novelty adaptation. 
In online \textit{continuous} or \textit{lifelong learning}~\citep{silver2013lifelong} and \textit{online learning}~\citep{shalev2012online,hazan2016introduction}, 
a learner makes the assumption that the world is too complex or unpredictable to learn offline, functionally treating the world as perpetually novel.
Instead of making a training-deployment distinction, these learners continuously update.
With these techniques, there may be little-to-no guarantee of convergence pre- or post-novelty. 

When starting with some pre-novelty knowledge, even if the naive online learning approach results in post-novelty convergence, this could induce catastrophic inference~\citep{mccloskey1989catastrophic}, causing the agent to transfer little, if any, of its previous model. 
This forgetting phenomenon can make learning less efficient. 
\textit{Transfer learning}~\citep{zhu2021transfer} seeks to overcome this limitation by transferring the model to an environment that differs in a small but non-trivial way.
While this approach has similarities to novelty adaptation, transfer learning techniques often assume that an agent can return to an offline training phase when encountering the new dataset or environment, whereas novelty adaptation must occur during deployment. 



\vspace{-0.5\baselineskip}
\paragraph{\bf World Models in Reinforcement Learning.}
Reinforcement learning can be broadly categorized as \textit{model-based} and \textit{model-free} \citep{sutton2018reinforcement}.
In model-based reinforcement learning, an agent learns a model of the \textit{transition function} and uses it to formulate a policy with an iterative technique such as Monte Carlo tree search \citep{silver2017mastering} or a dynamic programming approach such as value iteration. 
In model-free reinforcement learning, a model of the policy is learned directly, without the need for a transition function, and then sampled to determine the actions of an agent. 
For novelty adaptation, model-based methods may be more sample-efficient, however, they tend to be more sensitive to modeling error than model-free methods \citep{sutton2018reinforcement}.

\textit{World models} learn both the transition function and the policy, using them together to drive agent performance \citep{ha_recurrent_2018, schmidhuber1990line}. 
DreamerV2~\citep{hafner2020mastering} represents the state of the art in world model-based reinforcement learning. 
In the Dreamer approach, 
only the world model learns directly from interacting with the environment, while the policy is trained exclusively from interactions ``imagined'' by the world model~\citep{hafner2019dream}. 
This builds on prior world model research that used imagination to help train standard RL models \citep{racaniere_imagination_2017, ha_recurrent_2018, lin2020improving} and multi-agent models \citep{pretorius2020learning}. 
DreamerV2 is more sample-efficient than most model-free reinforcement learning agents but still requires a very large model; it was not explicitly designed for novelty-handling.
No prior work uses world model based RL for novelty adaptation.

Reinforcement learning typically models environments as standard Markov decision processes (MDP): $M=\langle S, A, R, P, \gamma\rangle$, 
where $S$ is the set of environment states, $A$ is the set of actions, $R: S \times A \rightarrow \mathbb{R}$ is the function mapping from states and actions to a scalar reward, $P: S \times A \rightarrow \mathbb{S}$ is the transition function between states, and $\gamma$ is the discount factor. 
The injection of novelty, after an arbitrary and {\em a priori} unknown number of episodes or games $t$ constitutes a transformation from the original environment or MDP $M$ to a new environment or MDP $M'$. 
All aspects of the problem except an agent's decision-making model are considered properties of the environment. 
This includes agent morphology, action preconditions, and effects. 
%
We constrain the problem in a few ways. We assume that each agent's observation space and action space remain consistent throughout. 
We also assume that the agent's mission $T$, and therefore the extrinsic reward function, is consistent throughout.

%% file: 4_method.tex
\sysname{} is an end-to-end trainable neuro-symbolic world model comprised of two components: (1) a neural policy and (2) a symbolic world model. 
The symbolic world model (which we also refer to as the ``rule model'') consists of {\em rules} that, in aggregate, approximate the environment's latent transition function. 
The rule model serves two core functions. 
First, the rule model learns to predict state transitions pre-novelty. 
Rule violations  thus indicate the introduction of novelty and the need to update the rule model and the policy.
Second, once in a post-novelty environment, \sysname{} uses the rule model to simulate the environment, enabling rollouts for retraining the neural policy model so as to require fewer interactions with the real environment. 
Shown in Figure~\ref{fig:architecture}, this interaction between the world model and the policy allows \sysname{} to trust its policy pre-novelty, then depend more heavily on its world model post-novelty.  
Our world model algorithm is designed so that the rule model is independent of the neural policy implementation, making our approach compatible with any policy framework that uses the same data inputs as the rule model.
For our implementation of \sysname{}, we use Proximal Policy Optimization (PPO)~\citep{schulman2017proximal} on an Advantage Actor-Critic (A2C) neural architecture. 

\subsection{Interval-Based Symbolic World Model}

In \sysname{} the symbolic world model--modeling the transition function $P$--is represented as a set of $K$ rules $\{\rho_k\}$ of the form $\langle c_s, c_a, e\rangle$. 
In this representation, $c_s$ is a state precondition, $c_a$ is the action precondition (similar to a do-calculus precondition \texttt{do(a)}), and $e$ is an effect. 
A rule $\rho$ is determined to apply if the input state $s$ and action $a$ match that rule's preconditions.
The state preconditions contain a set of values corresponding to a subset of state features $\phi_1...\phi_m$. 
When both the state and action preconditions of $\{\rho\}$ are satisfied, then $\{\rho\}$ is applicable and can be executed if chosen. 
Effects $e$ are the difference between the input state and the predicted state: $e=s^{'}-s$.  
This formulation has similarities to logical calculus frameworks such as ADL and PDDL~\citep{McDermott2000The1A} by encoding preconditions and effects.
Our approach is designed to be learned, rather than engineered, similar to ``game rule'' learning~\citep{guzdial:ijcai2017}. 

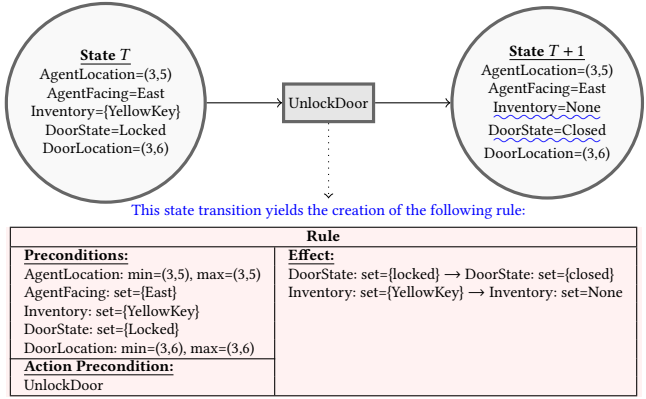
\begin{figure}
\centering
\scriptsize
    \begin{tikzpicture}[roundnode/.style={circle, draw=black!60, fill=gray!5, very thick, minimum size=7mm},
    squarenode/.style={rectangle, draw=black!60, fill=gray!20, very thick, minimum size=5mm}]
    \node[roundnode, align=center] (state1) {\underline{\bf State $T$}\\AgentLocation=(3,5)\\AgentFacing=East\\Inventory=\{YellowKey\}\\DoorState=Locked\\DoorLocation=(3,6)};
    \node[squarenode, align=center] (action) [right=of state1] {UnlockDoor};
    \node[roundnode, align=center] (state2) [right=of action]{\underline{\bf State $T+1$}\\AgentLocation=(3,5)\\AgentFacing=East\\\blueuwave{Inventory=None}\\\blueuwave{DoorState=Closed}\\DoorLocation=(3,6)};
    \node (txt) [below=of action] {\textcolor{blue}{This state transition yields the creation of the following rule:}};
    \draw[->] (state1.east) -- (action.west);
    \draw[->] (action.east) -- (state2.west);
    \draw[->, dotted] (action.south) -- (txt.north);
    \end{tikzpicture}
    
    \colorbox{rulecolor}{
    \begin{tabular}{|l|l|}
    \hline
    \multicolumn{2}{|c|}{\bf Rule}\\
    \hline
    \underline{\bf Preconditions:} & \underline{\bf Effect:} \\
    AgentLocation: min=(3,5), max=(3,5) & DoorState: set=\{locked\} $\rightarrow$ DoorState: set=\{closed\}\\
    AgentFacing: set=\{East\} & Inventory: set=\{YellowKey\} $\rightarrow$ Inventory: set=None \\
    Inventory: set=\{YellowKey\} &\\
    DoorState: set=\{Locked\} & \\
    DoorLocation: min=(3,6), max=(3,6) &\\
    \cline{1-1}
    \underline{\bf Action Precondition:} &\\ 
    UnlockDoor & \\
    \hline
    \end{tabular}
    }
\caption{Top shows example environmental states passed to the rule learner (changes underlined). Bottom shows the learned world model rule describing the key opening a door. }
\label{fig:rule-creation}
\end{figure}

\sysname{} uniquely formulates preconditions as a set of \textit{axis-aligned bounding intervals} (AABIs), also known as hyperrectangles or
$n$-orthotopes \citep{coxeter1973regular} in feature space that cover the training data.
AABIs are simple, $d$-dimensional convex geometries that, given a set of sample points to group $x_1...x_n$, define the minimum interval along each dimension that encloses the entire set.
Regardless of the size of the interval, AABIs can be defined by two $d$-dimensional points---a minimum and maximum bound---which makes them very efficient to query for both training and inference.
They can accommodate a mixture of continuous and categorical (non-continuous) variables, both of which are common in symbolic methods, where categorical AABI values are simply the exact set of matching values.
For example, see the bottom of Figure~\ref{fig:rule-creation}, which shows the AABIs for a rule for unlocking a door in the NovGrid grid world. In this case the intervals limit the action to a single agent location $(3,5)$. Figure~\ref{fig:rule-relaxation} shows another example where the action is applicable to an interval of locations.



The AABIs of the preconditions do not need to be intersecting; each unique rule can have multiple disjoint intervals. Our rule update algorithm (see next section) minimizes the number of different intervals.
Multiple rules per action will exist when actions have different effects depending on the current state.
For example, in a grid world, the \texttt{forward} action changes the agent's $\phi$
positional feature in the state along the direction the agent is facing when nothing is in the way, 
but \texttt{forward} will have no effect on the state if there is a wall directly in front of the agent. 
This same functionality enables us to account for probabilistic transitions; multiple rules will have the same action and state precondition but different \textit{effect distributions}.
Using the example of opening a locked door with a key, there is the possibility that a lock is ``sticky'' and an agent may require several tries. The effect of the rule that predicts the opening of the lock would then be a distribution over the 
rules with identical preconditions but different effects.


\begin{algorithm}[t]
\scriptsize
\SetAlgoLined
\KwIn{Prior State $s_{t-1}$, Action $a_{t-1}$, NextState $s$, WorldModel rule set $P$}
\KwOut{Applied Rule $\rho$}
StateChange $\delta s = s - s_{t-1}$\;
$RuleHit = \texttt{False}$\;
\For{Rule $\rho_k = \langle c_{s,k}, c_{a,k}, e_k\rangle$ in $P$}{
    \If{$a_{t-1} == c_{a,k}$}{
        \eIf{$\delta s == e_k$}{
            \eIf{$CollisionCheck(s_{t-1}, c_{s,k})$}{
                $RuleHit \leftarrow \texttt{True}$ \;
                return $\rho$ \;
            }
            {
                $RuleRelaxation(\rho, s_{t-1})$ \;
                $RuleHit \leftarrow \texttt{True}$ \;

            }
        }
        {
            \eIf{$CollisionCheck(s_{t-1}, c_{s,k})$}{
                $RuleCollisionResolution(\rho, s_{t-1})$ \;
            }
            {
                $RuleCreation(c_{s} = s_{t-1}, c_{a} = a_{t-1}, e = \delta s)$ \;
                $RuleHit \leftarrow \texttt{True}$ \;

            }
        }
    }
}
\If{$RuleHit == \texttt{False}$}{
    $RuleCreation(c_{s} = s_{t-1}, c_{a} = a_{t-1}, e = \delta s)$ \;
}
\caption{Rule Model Update}
\label{alg:update}
\end{algorithm}

\subsection{Rule Learning}

The rule learning process (Algorithm~\ref{alg:update}) constructs a compact, collision-free set of rules that provide maximum coverage of the state-action space while minimizing the complexity of the symbolic world model.
Moreover, it is an online updating process; once a rule is learned, it can be updated without knowledge of past observations. 

The rule update process begins with the rule model initialized as an empty set.
After an action is taken in the environment, the rule learner receives the prior state of the environment from which to derive a precondition, 
the action taken, and a new state of the environment from which to derive an effect. 
Comparing the prior state, action, and new state with the state preconditions, action preconditions, and effects (respectively) of any existing rules in the model, 
the update algorithm enters one of four cases: 
\begin{enumerate}
    \item {\em No Change}: The prior state falls inside the state precondition AABI of an existing rule with a matching action and effect.
    \item \textit{Rule Creation}: There is no rule where the action precondition is satisfied or the state difference matches the effect.
    A new ``point'' rule is created that exactly describes the prior state.
    \item \textit{Rule Relaxation}: A rule exists where the action precondition is satisfied and state difference matches the effect, but the prior state is not covered by the existing rule's state precondition AABI. 
    The rule is ``relaxed'' by expanding the AABI.
    \item \textit{Rule Collision Resolution}: A rule exists where the action precondition and state precondition AABI are satisfied but the effect is different. 
    The AABI of the existing rule is split with a minimum cut (min-cut) operation. 
\end{enumerate}

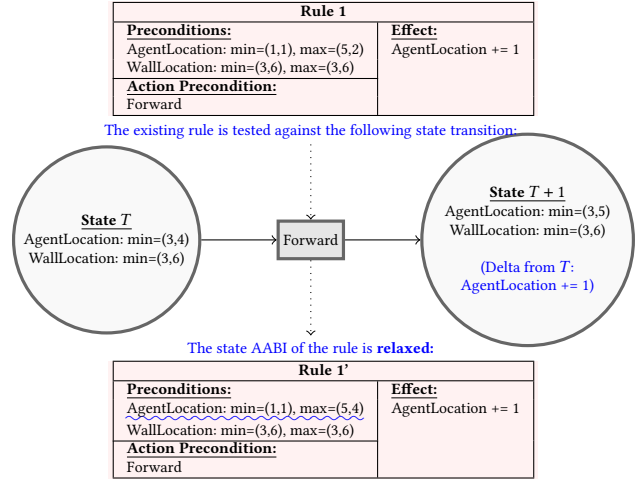
\begin{figure}
\centering
\scriptsize
    \colorbox{rulecolor}{
    \begin{tabular}{|l|l|}
    \hline
    \multicolumn{2}{|c|}{\bf Rule 1}\\
    \hline
    \underline{\bf Preconditions:} & \underline{\bf Effect:} \\
    AgentLocation: min=(1,1), max=(5,2) & AgentLocation += 1\\
    WallLocation: min=(3,6), max=(3,6) & \\
    \cline{1-1}
    \underline{\bf Action Precondition:} &\\ 
    Forward & \\
    \hline
    \end{tabular}
    }
    
    \begin{tikzpicture}[roundnode/.style={circle, draw=black!60, fill=gray!5, very thick, minimum size=7mm},
    squarenode/.style={rectangle, draw=black!60, fill=gray!20, very thick, minimum size=5mm}]
    \node[roundnode, align=center] (state1) {\underline{\bf State $T$}\\AgentLocation: min=(3,4)\\WallLocation: min=(3,6)};
    \node[squarenode, align=center] (action) [right=of state1] {Forward};
    \node[roundnode, align=center] (state2) [right=of action]{\underline{\bf State $T+1$}\\AgentLocation: min=(3,5)\\WallLocation: min=(3,6)\\\\\textcolor{blue}{(Delta from $T$:}\\\textcolor{blue}{AgentLocation += 1)}};
    \node (txt1) [above=of action] {\textcolor{blue}{The existing rule is tested against the following state transition:}};
    \node (txt2) [below=of action] {\textcolor{blue}{The state AABI of the rule is \textbf{relaxed:}}};
    \draw[->] (state1.east) -- (action.west);
    \draw[->] (action.east) -- (state2.west);
    \draw[->, dotted] (txt1.south) -- (action.north);
    \draw[->, dotted] (action.south) -- (txt2.north);
    \end{tikzpicture}
    
    \colorbox{rulecolor}{
    \begin{tabular}{|l|l|}
    \hline
    \multicolumn{2}{|c|}{\bf Rule 1'}\\
    \hline
    \underline{\bf Preconditions:} & \underline{\bf Effect:} \\
    \blueuwave{AgentLocation: min=(1,1), max=(5,4)} & AgentLocation += 1\\
    WallLocation: min=(3,6), max=(3,6) & \\
    \cline{1-1}
    \underline{\bf Action Precondition:} &\\ 
    Forward & \\
    \hline
    \end{tabular}
    }
    
\caption{Rule Relaxation example, where the blue underlined precondition AABI corresponding to the agent location has been expanded in the modified Rule 1' 
to include agent location from state S.}
\label{fig:rule-relaxation}
\end{figure}



When the agent takes an action, the rule learner observes the prior state at time $T$, the action executed, and the next state at time $T+1$.
If rules exist where the state transition satisfies the action precondition, the state precondition, or the effect, we first try to modify existing rules using Rule Relaxation and Rule Collision Resolution. 
Rule Creation is only necessary if collision resolution and relaxation do not apply.
Given the agent's performed action, we initially only consider rules where the action precondition is satisfied. 
We then identify whether the state prior to the action is contained in any of these existing rules' AABI.
This is achieved using the geometric hyperplane separation theorem (also called the separating axis theorem)~\citep{hastie2009elements, ball1997elementary}. 
Geometrically, we can assert that for all features $\phi_d \in \Phi$, if there exists a feature for which the point is less than the min or more than the max of an interval, then a separating hyperplane exists.
The theorem states simply that if a hyperplane exists in feature space between the point and the geometric shape, there is no collision and the point is outside the AABI. 
Specifically, for prior state $s_{t-1}$ and an AABI $I = [I_{min},I_{max}]$:

\begin{equation}
    s_{t-1} \not\in I \iff \exists \phi~s.t.~[ s_{t-1,\phi} > I_{max,\phi}  \cup  s_{t-1,\phi} < I_{min,\phi} ].
\end{equation}

\textbf{Rule Creation}
is straightforward: the combination of prior state and action in an observed state-action-state transition does not fall within the AABI of any existing rule.
A new rule is added to the rule model where the AABI for continuous state features are assigned \texttt{min} and \texttt{max} values equal to their current value, and categorical state feature values are singleton members of their features. 
Similarly, the rule's action precondition is set to the action in the state transition, and the effect is equal to the difference between $s_{t-1}$ and $s$. 
This process is illustrated in Figure \ref{fig:rule-creation}.
We describe a rule that is circumscribed to a single set of values for each feature as a``point'' rule. 

\textbf{Rule Relaxation} seeks to expand an AABI of an existing rule to match the newly encountered action precondition and effect. 
The operation is simple: for the prior state $s_{t-1}$ and AABI $I$ represented by points $I_{min}$ and $I_{max}$, the relaxed min and max points are
$I^*_{min} = \texttt{min}(s_{t-1},I_{min})$ and $I^*_{max} = \texttt{max}(s_{t-1},I_{max})$.
Figure \ref{fig:rule-relaxation} illustrates rule relaxation.
Here, a previously learned rule exists that models the change in agent position caused by a \texttt{forward} action.
This rule matches the action precondition and effect but not the state precondition, possibly because it simply had not been observed yet in that part of the environment. 
As a result, the AgentPosition AABI is expanded to include the observation.

After this expansion, the new AABI $I^*$ is checked once more, this time for collision with other existing rules, using the same geometric hyperplane separation theorem described above.
For comparing intervals, however, we check for a hyperplane between $I*$ and the AABI $I^k$ of another rule $\rho_k$ by comparing the maxima to the minima of the intervals.
Given maximum and minimum points $[I^*_{min},I^*_{max}]$ and $[I^k_{min},I^k_{max}]$, 
if there exists a feature $\phi$ for which $I^*_{min} > I^k_{max}$, or vice versa, then there is no collision. 
If a collision does exist, instead of trying to compromise between the two rules, we execute Rule Collision Resolution.

\begin{figure}
\centering
\scriptsize
    \colorbox{rulecolor}{
    \begin{tabular}{|l|l|}
    \hline
    \multicolumn{2}{|c|}{\bf Rule 1}\\
    \hline
    \underline{\bf Preconditions:} & \underline{\bf Effect:} \\
    AgentLocation: min=(1,1), max=(8,8) & AgentLocation += 1\\
    WallLocation: min=(3,6), max=(3,6) & \\
    \cline{1-1}
    \underline{\bf Action Precondition:} &\\ 
    Forward & \\
    \hline
    \end{tabular}
    }
    
    \begin{tikzpicture}[roundnode/.style={circle, draw=black!60, fill=gray!5, very thick, minimum size=7mm},
    squarenode/.style={rectangle, draw=black!60, fill=gray!20, very thick, minimum size=5mm}]
    \node[roundnode, align=center] (state1) {\underline{\bf State $T$}\\AgentLocation: min=(3,5)\\WallLocation: min=(3,6)};
    \node[squarenode, align=center] (action) [right=of state1] {Forward};
    \node[roundnode, align=center] (state2) [right=of action]{\underline{\bf State $T+1$}\\AgentLocation: min=(3,5)\\WallLocation: min=(3,6)\\\\\textcolor{blue}{(Delta from $T$: None)}};
    \node (txt1) [above=of action] {\textcolor{blue}{The existing rule is tested against the following state transition causing a \textbf{collision}:}};
    \node (txt2) [below=of action] {\textcolor{blue}{The Rule is \textbf{split} about the AgentLocation state precondition:}};
    \draw[->] (state1.east) -- (action.west);
    \draw[->] (action.east) -- (state2.west);
    \draw[->, dotted] (txt1.south) -- (action.north);
    \draw[->, dotted] (action.south) -- (txt2.north);
    \end{tikzpicture}
    
    \colorbox{rulecolor}{
    \begin{tabular}{|l|l|}
    \hline
    \multicolumn{2}{|c|}{\bf Split Rule 1'}\\
    \hline
    \underline{\bf Preconditions:} & \underline{\bf Effect:} \\
    \blueuwave{AgentLocation; min=(1,1), max=(8,5)} & AgentLocation += 1\\
    WallLocation: min=(3,6), max=(3,6) & \\
    \cline{1-1}
    \underline{\bf Action Precondition:} &\\ 
    Forward & \\
    \hline
    \end{tabular}
    }
    
    \colorbox{rulecolor}{
    \begin{tabular}{|l|l|}
    \hline
    \multicolumn{2}{|c|}{\bf Split Rule 1''}\\
    \hline
    \underline{\bf Preconditions:} & \underline{\bf Effect:} \\
    \blueuwave{AgentLocation; min=(1,6), max=(8,8)} & AgentLocation += 1\\
    WallLocation: min=(3,6), max=(3,6) & \\
    \cline{1-1}
    \underline{\bf Action Precondition:} &\\ 
    Forward & \\
    \hline
    \end{tabular}
    }
    
    \textcolor{blue}{And a new rule is \textbf{created}:}
     
    \colorbox{rulecolor}{
    \begin{tabular}{|l|l|}
    \hline
    \multicolumn{2}{|c|}{\bf New Rule 2}\\
    \hline
    \underline{\bf Preconditions:} & \underline{\bf Effect:} \\
    \blueuwave{AgentLocation; min=(3,5), max=(3,5)} & \blueuwave{None}\\
    \blueuwave{WallLocation: min=(3,6), max=(3,6)} & \\
    \cline{1-1}
    \underline{\bf Action Precondition:} &\\ 
    \blueuwave{Forward} & \\
    \hline
    \end{tabular}
    }
\caption{\textit{Rule Collision}, and the resulting rule \textit{split} and \textit{creation}. The blue-underlined preconditions in the newly split Rule 1' and Rule 1'' indicate the feature dimension along which the original Rule 1 is split. The newly created Rule 2 accounts for the state transition that caused the collision with the original Rule 1.}
\label{fig:rule-collision}
\end{figure}
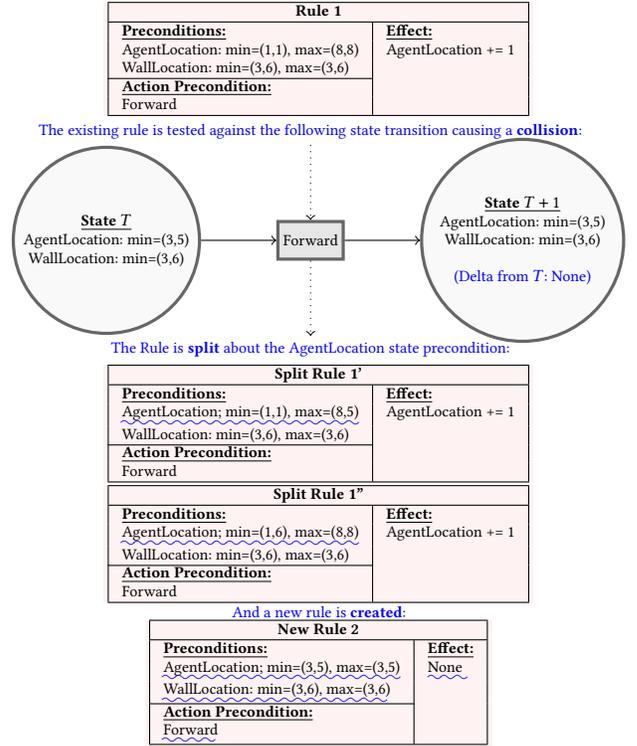

\textbf{Rule Collision Resolution} finds the min-cut partition of the existing rule's state precondition AABI. 
Consider the AABI as a graph, where the graph nodes are the bounding hyperplanes (``faces'' of the hyperrectangle) and the edges are lines that connect opposing hyperplanes weighted by their length. 
This min-cut of the AABI is simply the division along the largest feature axis that intersects with the prior state. 
The new, divided AABIs are added to the existing rule's preconditions, the original AABI is removed, and the prior state is assessed for accommodation with Rule Creation or Rule Relaxation because it may not be included in either split. 
Figure \ref{fig:rule-collision} illustrates a situation where the \texttt{forward} rule is observed to not correctly predict the outcome of a state transition \texttt{forward} because the agent hits a wall.
Because the rule describing this state transition has a precondition subsumed by the existing \texttt{forward} rule but with a different effect, there is a rule collision. 
The collision is resolved by splitting the rule. 
A new point rule can be created because the prior state is now outside the AABIs of the split rules.

Post-novelty, splitting rules in Rule Collision Resolution can result in one of the split rules having a precondition with a feature with empty interval or empty categorical set.
If this occurs, the ``empty split'' is discarded. 





\subsection{Novelty Detection}

Once pre-novelty neural policies converge and the symbolic world model is created as described above, learning is turned off for both; this saves compute and allows the policy to focus on exploitation. 
As the agent performs tasks in the environment, it looks for state-action-state transitions that are inconsistent with the rules in the world model and triggers adaptation when one of two cases occur.
(1)~The same rule is violated $n$ consecutive times.
A violation is defined as the observed subsequent state of the environment not matching a rule's expected state effect, though the state and action preconditions both match.
(2)~The same observed state causes a failed prediction in more than $n$ consecutive visits.
$n$ is a hyperparameter for sensitivity to novelty, set to $n=2$ in our experiments.

Once a novelty is detected, 
the neural policy and the symbolic world model begin to update online again. 
The post-novelty rule update process is exactly the same as pre-novelty rule learning (Algorithm~\ref{alg:update}).
The rule model can thus be updated with as little as a single iteration of the rule learning algorithm, with guaranteed improved next-state prediction.


\begin{algorithm}[t]
\scriptsize
\SetAlgoLined
\KwIn{Pre-Novelty Policy $\pi$, World Model $P$, Mix Ratio $\eta$}
$PostNov \leftarrow False$\;
$s_t \leftarrow$ initial observation\;
\While{$true$}
{
Select action $a_{t}=\pi(s_{t})$ \;

Predict next state $\hat{s}_{t+1} = P(s_{t}, a_{t})$ \;

Execute $a_{t}$ in $Env$ and observe 
next state and reward 
$s_{t+1}, r_{t}$\;

$PostNov \leftarrow [PostNov$ {\bf OR} $DetectNovelty(P,\hat{s}_{t+1}, s_{t+1})$] \;

\If{$PostNov$}
{
Add $\langle s_t, a_t, s_{t+1}, r_t\rangle$ to $UpdateBuffer$\;

$P \leftarrow RuleModelUpdate(s_t, a_t, s_{t+1})$\;

Add $ImagineRollouts(P)$ per $\eta$ to $UpdateBuffer$\;

Periodically update $\pi$ with $UpdateBuffer$ \; 
}
$s_t \leftarrow s_{t+1}$
}

\caption{Imagination-Based Adaptation}
\label{alg:imagination-based adaptation}
\end{algorithm}


\begin{table*}[t!]
\footnotesize
\centering
\begin{tabular}{c|c|c|c|c|}
{} & {\bf Adaptive Efficiency } & {\bf Pre-novelty} & {\bf Asymptotic } &  {\bf Update Efficiency}\\
& {\bf @0.95 (steps) $\downarrow$} & {\bf Performance} $\uparrow$ & {\bf Performance} $\uparrow$ &   {\bf (policy updates) $\downarrow$}\\
\hline
\multicolumn{5}{c}{\texttt{DoorKeyChange} novelty}\\
\hline PPO & 2.25E6 & 0.973 & 0.971 & 2.25E6 \\
\hline
 DreamerV2 & 5.3E5 & 0.971 & 0.973 &  3.82E8 \\
\hline
 \textbf{Ours} & 9.8E5 & 0.972 & 0.970 & \textbf{1.63E6} \\
\hline
\multicolumn{5}{c}{\texttt{LavaProof} novelty}\\
\hline
PPO & 1.39E5 & 0.972 & 0.991 & 1.39E5 \\
\hline
 DreamerV2 & Failed to adapt & 0.965 & Failed to adapt & Failed to adapt\\
\hline
 \textbf{Ours} & 8.3E4 & 0.972 & 0.991 & \textbf{1.38E5} \\
\hline
\multicolumn{5}{c}{\texttt{LavaHurts} novelty}\\
\hline
PPO & 2.08E6 & 0.992 & 0.971 & 2.08E6 \\
\hline
 DreamerV2 & 1.05E6 & 0.992 & 0.968 & 7.56E8 \\
\hline
 \textbf{Ours} & 1.07E6 & 0.992 & 0.972 & \textbf{1.78E6} \\
\hline
\end{tabular}
  \caption{Novelty metric results
  averaged over three runs.
  DreamerV2 did not adapt to the novelty on \texttt{LavaProof}.
}
\label{table:main}
\end{table*}

\subsection{Imagination-Based Policy Adaptation}

Post-novelty, the newly updated rules within the symbolic world model reflect the agent's belief about the post-novelty state transition function. 
The agent now uses that rule model to ``imagine'' how sequences of actions will play out---which we refer to as ``imagination-based simulation''---and update its policy without interacting or executing actions in the true environment.
Specifically, as we show in Algorithm \ref{alg:imagination-based adaptation},
we use the rule model to simulate state-action-state transitions that then populate the agent's update buffer--the data on which the policy will be trained.
The policy training algorithm generates a loss over samples drawn from the update buffer and back-propagates loss through the policy model.
Figure~\ref{fig:architecture} shows how the standard Actor-Critic neural architecture and imagination-based simulation work together to feed real and imagined state observations.

The agent follows its policy in the imagined environment and repeatedly experiences the first rule change's consequences, receiving a reduced (or increased) expected reward, pushing the policy away from (or toward) the impacted actions.
As the symbolic world model detects new discrepancies and represents the post-novelty environment more accurately, the policy may be able to ``imagine'' experiencing and accommodating novelty with minimal exposure to the novelty in the environment. 

Continuing the example of the re-keyed door lock (Figure~\ref{fig:splash}, top), the agent has executed its pre-novelty policy of navigating to the yellow key and then to the yellow door only to discover that the door no longer opens. 
Upon arriving at the state $\langle$\textit{AgentInFrontOfDoor,  AgentCarryingYellowKey, DoorLocked}$\rangle$ 
and performing the action \texttt{unlock}, the agent expects $DoorUnlocked$ to become true.
However, since the door has been re-keyed the \texttt{unlock} action results in no change. 
When this occurs, there is a rule collision, resolved by creating a rule with a $\{\emptyset\}$ effect delta, and the old rule is ``split''. In this case, the split results in states with empty AABIs, in which case the rule is deleted. 
The rule collision is detected as a novelty, and the agent begins updating its policy.
Specifically, the agent no longer receives the utility of walking through the door and the policy updates to reflect reduced utility of being directly in front of the door. 
Over time, the utility of the state will decrease enough that the agent will prefer other states, increasing exploration and eventually coming across the blue key.
This switch from exploitation to exploration is accelerated by the agent's ability to repeatedly imagine arriving before the door and choosing alternative actions because there is no valid rule in which the door opens.

\sloppypar{Eventually through exploration the agent will find itself before the door with the blue key: $\langle$\textit{AgentInFrontOfDoor, \blueuwave{AgentCarryingBlueKey}, DoorLocked}$\rangle$ 
Trying to open the door with the blue key will this time result in the effect of $DoorUnlocked$ changing to $DoorClosed$.
This, in conjunction with the \texttt{unlock} action precondition does not match any existing rule, and a new rule is created, at which point imagination will facilitate faster policy learning.
Once again having access to the utility and reward of states beyond the door, the agent will converge to a new policy involving the blue key.
}

To account for world model error, our policy is also trained on real environment interactions.
A mixing ratio parameter controls the ratio of imagined and real environment rollouts in the policy update buffer;
For every $t$ steps in the real environment, the agent runs $\frac{t}{\eta}$ steps in the imagined environment. 
This mixing effectively adds noise to the policy update and
helps drive the policy back into an ``explore'' mode, where actions will be selected more randomly by the policy.

%% file: 5_experiments.tex
We perform experiments in the NovGrid~\citep{balloch2022novgrid}
environment, which extends the MiniGrid~\citep{gym_minigrid} environment with novelties that can be injected at an arbitrary time during testing. 
NovGrid allows for controlled, replicable experiments with stock novelties. 
We use two 8x8 MiniGrid environments: 
(1)~\texttt{DoorKey} where an agent must pick up a key, unlock a door, and navigate to the goal behind that door, and 
(2)~\texttt{LavaShortcutMaze} where an agent must navigate a maze that has a pool of lava lining the side of the maze nearest to the goal.
In all cases we used the default sparse MiniGrid reward.

We tested the performance of our method on three novelty types:
\textit{shortcut}, \textit{delta}, and \textit{barrier} \citep{balloch2022novgrid}.
\texttt{LavaProof} is a shortcut novelty that makes lava in the \texttt{LavaShortcutMaze} environment to be harmless to the agent (where pre-novelty it destroyed the agent), offering a shorter path to the goal. 
\texttt{DoorKeyChange} is a delta novelty that changes which of two keys unlock a door in the \texttt{DoorKey} environment, 
not changing the difficulty of reaching the goal but requiring different state-action sequences of similar length and complexity.
\texttt{LavaHurts} is a barrier novelty and the inverse of \texttt{LavaProof}, 
changing the effect of lava in the \texttt{LavaShortcutMaze} to destroy the agent (where pre-novelty, lava was harmless), thereby eliminating the shorter lava path to the goal. 
%
The \texttt{DoorKeyChange} and \texttt{LavaProof} novelties are conceptually shown in Figure~\ref{fig:splash}.

\begin{figure}[t]
\centering
\includegraphics[width=0.925\linewidth]{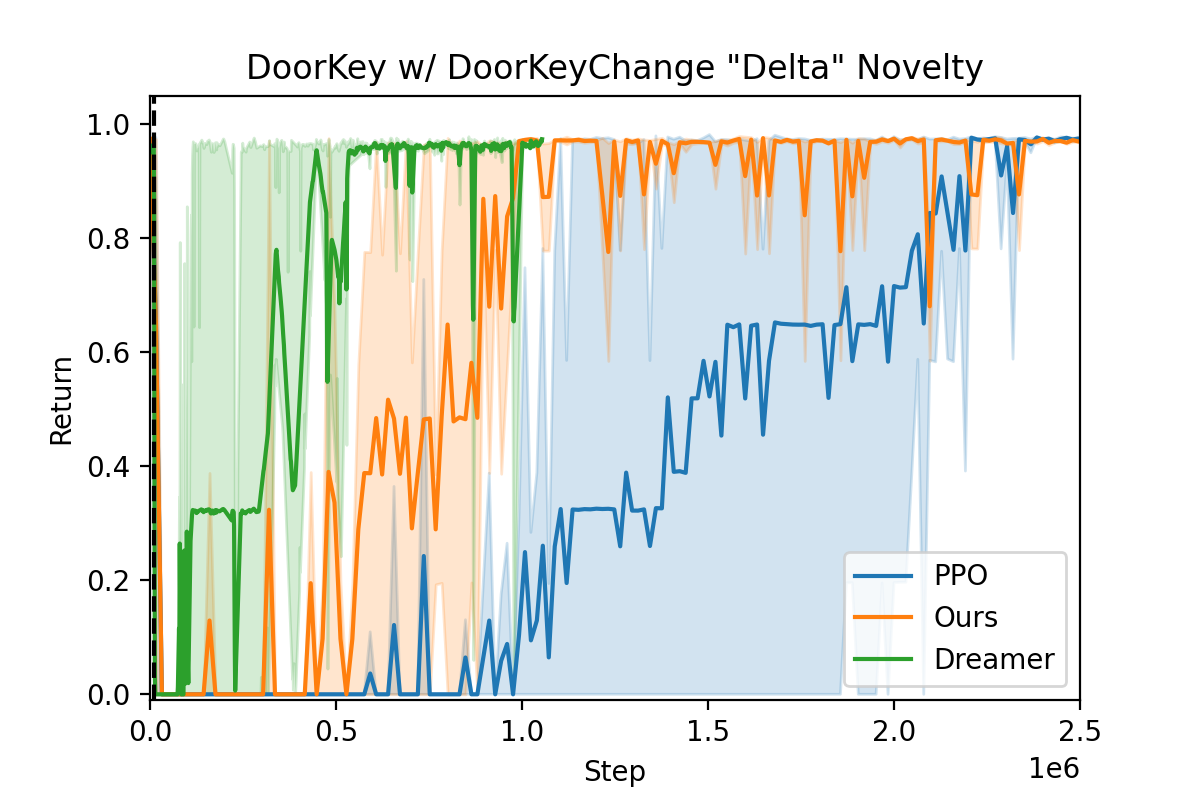}
\includegraphics[width=0.925\linewidth]{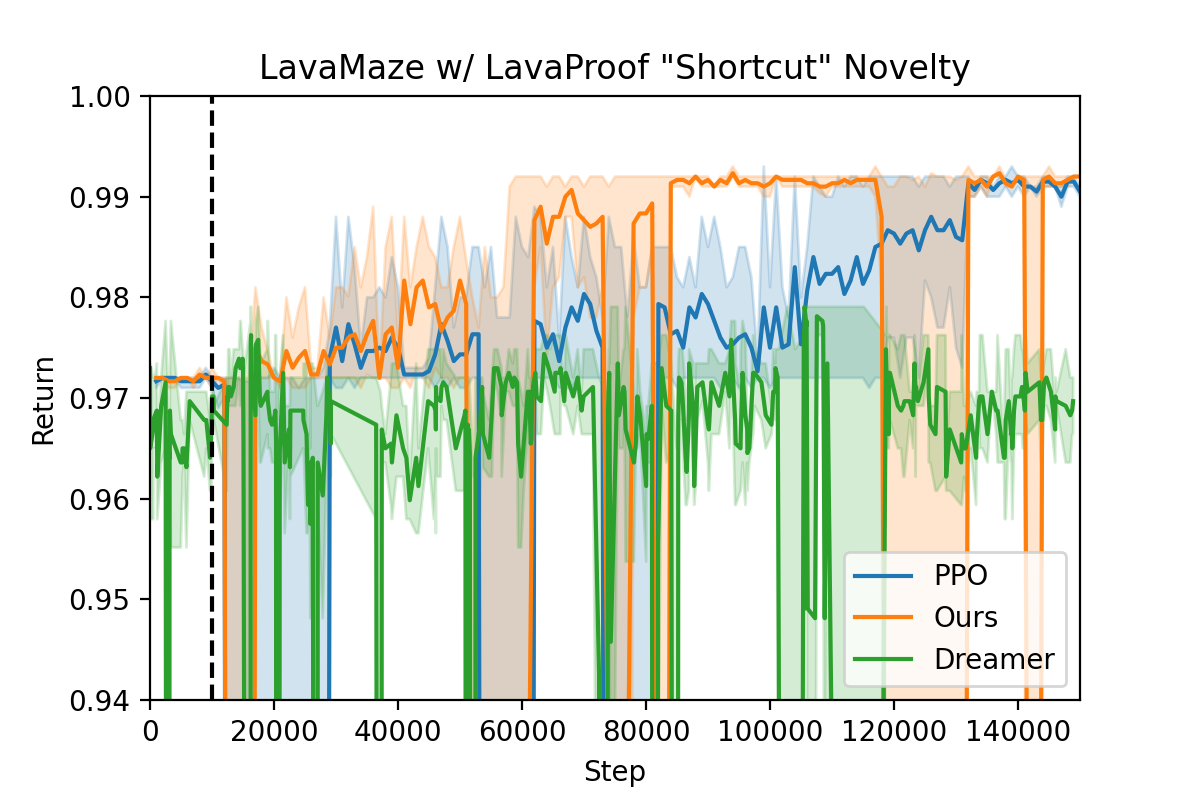}
\includegraphics[width=0.925\linewidth]{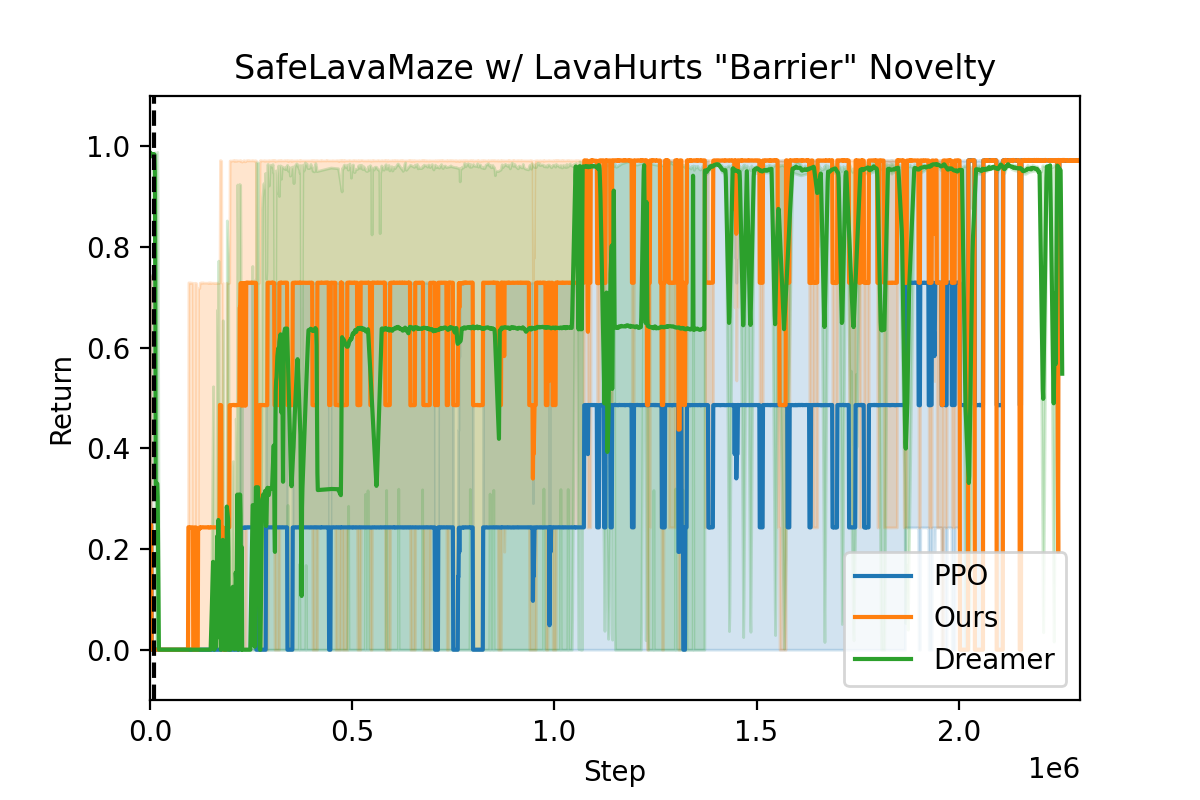}
\caption{
These plots show the adaptive performance post-novelty. 
Each method experiences novelty injection at 10,000 environment steps (signified by the vertical dotted black line). 
{\bf Top} shows novelty response for the \texttt{DoorKeyChange} ``delta'' novelty on the \texttt{DoorKey} environment. 
{\bf Middle} shows novelty response for the \texttt{LavaProof} ``shortcut'' novelty in the \texttt{LavaShortcutMaze} environment. Note: Dreamer does not find the shortcut. 
{\bf Bottom} shows novelty response for the \texttt{LavaHurts} ``barrier'' novelty in the \texttt{LavaShortcutMaze} environment, where lava is safe for the agent pre-novelty. 
}
\label{fig:env_steps}

\end{figure}

To evaluate performance in these test environments, we adopt two metrics for novelty adaptation from \citet{balloch2022novgrid} that builds on \citet{alspector2021representation}. 
(1) {\em Asymptotic adaptive performance} is the final performance of the agent post-novelty relative to a random baseline, where higher is better. 
This is used to observe whether a method fully adapted to the post-novelty environment. 
(2) {\em Adaptive efficiency} is the number of time steps in the real, post-novelty environment required to converge to asymptotic adaptive performance, where fewer steps is better. 
In our work, this is achieved when the 10-step moving average method reaches $95\%$ of asymptotic adaptive performance. 
We add a third measure, {\em update efficiency}, which is the number of policy updates required post-novelty to reach asymptotic adaptive performance, where--again--fewer steps is better.

We compare \sysname{} post-novelty performance with two baselines.
The first is a standard reinforcement learning agent using Proximal Policy Optimization (PPO)~\citep{schulman2017proximal}, the same model-free reinforcement learning approach used by the neural policy in our~\sysname{} method.
The second baseline is DreamerV2~\citep{hafner2020mastering}, a state-of-the-art world modeling agent that learns an end-to-end neural world model.
The agents were not given any knowledge about the novelties at training time. 
All methods were allowed to train for as many as 10 million time steps to ensure convergence, and results for each method were averaged over three runs.
Novelty was injected at episode 50k, well after all agents had converged. 
Since the baseline agents lack novelty detection capabilities, we keep their learning on during evaluations so that agents can react immediately after novelty is injected.

The policy architectures for all agents use a convolutional neural net feature extractor and two fully connected output networks, one to estimate the value and one to serve as the policy functions of the agent.
All hyperparameters and architectures for DreamerV2 were consistent with the original publication~\cite{hafner2020mastering}, 
and all PPO hyperparameters are consistent with MiniGrid-suggested hyperparameters~\citep{willems_2020}.
For \sysname{} we use a mixing ratio of 60\% real rollouts to 40\% imagined rollouts.
This ratio was determined empirically by looking at the trade-off between asymptotic adaptive performance and adaptive efficiency.
At higher amounts of imagination, the agent did not recover full post-novelty performance.
The \sysname{} world model uses symbolic features from MiniGrid for rule learning, including the object type, color, and position of the agent and objects, the agent orientation, the agent's inventory, and whether doors are locked, unlocked, or open. 

%% file: 6_results.tex
\section{Results}

We document the results of these evaluations in Table \ref{table:main}, with the adaptation process of our method further illustrated in Figure \ref{fig:env_steps}. 
The table shows that pre-novelty, as expected, all three methods converge in all three novelty scenarios to effectively the same performance.
This means that all methods were able to find solution sequences of equal length to the goal for all environments. 

For the \texttt{DoorKeyChange} ``delta'' novelty, DreamerV2 slightly outperforms \sysname{} in adaptive efficiency, while both dramatically outperform PPO. 
From this result we can observe that imagination is strongly beneficial to post-novelty adaptation.
However, adaptive efficiency doesn't tell the entire story. 
DreamerV2 updates its policy on imagination only, which means that for each update to the world model, DreamerV2 must update its policy using many imagination iterations. 
As a result, update efficiency of DreamerV2 demands nearly two orders of magnitude more policy updates than \sysname{}. 

For the \texttt{LavaProof} ``shortcut'' novelty, \sysname{} substantially outperforms PPO in adaptive efficiency while DreamerV2 never finds the shortcut novelty.
Dreamer was trained for 2.5 million post-novelty environment steps in multiple runs with the same result.
We attribute DreamerV2's failure to the unique 
way in which its policy learner
depends on the accuracy of its world model.
Unfortunately, the world model continues to predict negative consequence of the lava.
Since the longer, pre-novelty solution still reaches the goal, Dreamer's world model remains fixed, imagination never varies, and the policy never updates. 
As a result, the world model can overfit easily if the policy produces the same sequence of actions every time.
PPO and \sysname{} do not encounter this issue because small variations in the policy occur that cause policy updates, which allows for more sensitivity to the shortcut novelty. 
However, these methods aren't perfect either; as can be seen in Figure \ref{fig:env_steps}.
It takes both methods more than 25,000 environment steps to react to the shortcut novelty injection, unlike in the barrier and delta novelties. 
This demonstrates a unique challenge in shortcut novelties: when the the pre-novelty optimal path still exists, converged, non-exploring agents will struggle to detect novelty.
As noted by \cite{balloch2022role}, future research is needed on novelty-aware exploration techniques. 

\begin{figure}
	\centering
\includegraphics[width=0.99\linewidth]{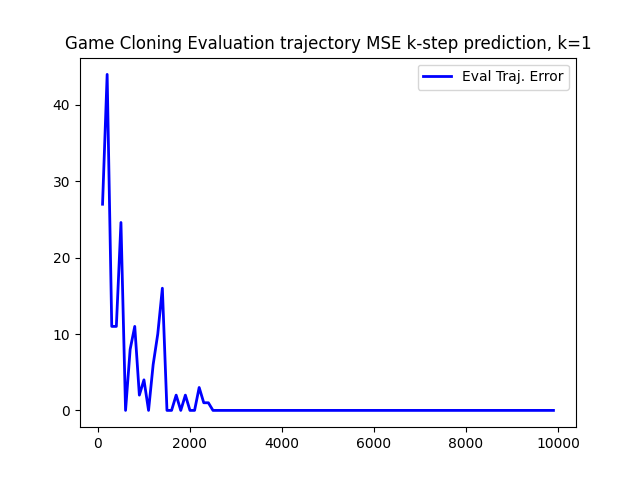}
\caption{This shows the \sysname{} 1-step prediction error vs environment steps during rule learner training in the Empty MiniGrid Environment.
}
\label{fig:cloner_steps}
\end{figure}

\sloppypar{
Finally, the \texttt{LavaHurts} ``barrier'' novelty is similar to the 
\texttt{DoorKeyChange} novelty, where \sysname{} and DreamerV2 have very similar adaptive efficiency both outperforming PPO by a wide margin, but in all cases taking a long time to converge.
What's more, in the \texttt{LavaHurts} and \texttt{DoorKeyChange} novelty, the high variance in both DreamerV2 and \sysname{} shows that in some of the trials both of these methods found the new solution very early in the adaption process, but could not consistently converge to it. 
Transfer learning tells us that both of these results are expected, as the simple pre-novelty solution does not prepare the policies for the more complex post-novelty environment. 

In general, our model iw able to make much more efficient use of our updates than DreamerV2. 
Because DreamerV2 only trains its policy in imagination, it runs hundreds of imagination policy updates for every world model update it executes. 
As shown in the last column of Table \ref{table:main}, blending the environment interaction data with the imagination data,~\sysname{} requires significantly fewer policy updates.
}

We also evaluate the efficiency and accuracy of the rule learner (results shown in Figure~\ref{fig:cloner_steps}). After every 100 training steps we would validate the model accuracy by running a random policy for 1000 steps and measure the average 1-step predictive error of the world model.
The pre-novelty rule learner requires only about 2000 training steps before it converges to near-perfect prediction accuracy. 
Grid worlds are simple, deterministic environments, so this is an unsurprising result.
The key is that it converges rapidly and is sample-efficient compared to neural world model learning techniques~\cite{hafner2019dream}.


These results show that \sysname{} with only 40\% imagination improves adaptation efficiency across all novelties over a neural policy with no world model. 
Furthermore, \sysname{} is competitive with DreamerV2 across these novelties and adapts with shortcuts that DreamerV2 fails to detect. 
Most interestingly, we can see from the last column of Table \ref{table:main} that \sysname{} achieves these results with fewer policy updates than PPO and DreamerV2.
This, however, will vary with the amount of imagination injected into the policy learner by \sysname{}, and should be the subject of further study.

%% file: 7_discussion_future.tex
As autonomous agents are deployed in open-world decision-making situations, techniques designed deliberately to handle novelty will be required.
This can include novelties from learning to unlock a door with a new key to discovering a new shortcut to reduce travel time or avoiding new hazards on previous safe solutions.

To this end, we show that reinforcement learning agent policies can be adapted more efficiently to novelties using symbolic world models that (1)~can be updated rapidly and (2)~simulate rollouts that then can be added to the policy learning process, thereby reducing the number of direct interactions with the post-novelty environment.
Specifically, our results show that \sysname{} is comparable in adaptation efficiency to state-of-the-art neural world modeling techniques while requiring only a fraction of policy updates. 
This suggests that symbolic representations are an effective complement to neural representations for adapting reinforcement learning agents to novelty.

For effective novelty adaptation, future techniques should consider how to leverage world models and their ability to learn models of the transition function and the policy together.
Our observations of world model-based agents adapting to novelty suggest that exploration is key to efficient novelty adaptation.
Even with access to world models, nearly all reinforcement learning agents assume that the converged pre-novelty policy can be followed greedily. 
This leads to a problematic feedback loop where policies trained with imagination are only able to access what the world model knows, which may not include the novelty, but these same policies direct the interaction with the environment. 
As a result, unless the novelty is along the path of the old optimal policy, the agent may never encounter the novelty, and the world model that trains it may never adapt.
New exploration techniques that operate in more plausible ``open-world'' environments will need to consider how to identify a novelty using the world model and how best to trade off this ``novelty exploration'' with following the agent's policy.